\documentclass[conference]{IEEEtran}
\usepackage{cite}
\usepackage{amsfonts}
\usepackage{textcomp}
\def\BibTeX{{\rm B\kern-.05em{\sc i\kern-.025em b}\kern-.08em
    T\kern-.1667em\lower.7ex\hbox{E}\kern-.125emX}}
    
\usepackage[final]{graphicx}
\graphicspath{ {./} }
\usepackage[reqno]{amsmath}
\usepackage{amssymb}
\usepackage{amsthm}
\usepackage{subfig}
\usepackage{epstopdf}
\usepackage{algorithm}
\usepackage{algorithmicx}
\usepackage{algpseudocode}
\usepackage{setspace}
\usepackage[T1]{fontenc}
\usepackage{color,soul}
\usepackage{multirow}
\usepackage{array}
\usepackage{cite}
\usepackage{verbatim}
\usepackage{enumitem}
\usepackage{url}
\usepackage{csvsimple}
\usepackage{float}

\usepackage{booktabs}   
\usepackage[table,xcdraw]{xcolor}  

\usepackage{graphicx}   
\usepackage{caption}    

\usepackage{float}
\usepackage{graphicx}
\usepackage[capitalize]{cleveref}
\usepackage{environ}
\usepackage{subfig}

\usepackage{rotating}
\usepackage{array,multirow,graphicx}

\usepackage{array}
\newcommand{\PreserveBackslash}[1]{\let\temp=\\#1\let\\=\temp}
\newcolumntype{C}[1]{>{\PreserveBackslash\centering}p{#1}}
\newcolumntype{R}[1]{>{\PreserveBackslash\raggedleft}p{#1}}
\newcolumntype{L}[1]{>{\PreserveBackslash\raggedright}p{#1}}

\definecolor{lightblue}{rgb}{0.678, 0.847, 0.902}  
\definecolor{mediumgray}{gray}{0.85}  
\definecolor{darkblue}{rgb}{0.0, 0.0, 0.6}
\definecolor{highlightgreen}{rgb}{0.5, 1.0, 0.5}  
\definecolor{mutedgreen}{rgb}{0.5, 0.7, 0.5}  

\begin{document}

\title{A Racing Dataset and Baseline Model for Track Detection in Autonomous Racing}

\author{\IEEEauthorblockN{\small Shreya Ghosh, Yi-Huan Chen, Ching-Hsiang Huang, Abu Shafin Mohammad Mahdee Jameel, Chien Chou Ho,  Aly El Gamal, Samuel Labi}
\IEEEauthorblockA{\small School of Electrical and Computer Engineering, Purdue University, USA}
\IEEEauthorblockA{\small {\{ghosh64, chen3966, huan1715, amahdeej, ho212, elgamala, labi\}@purdue.edu}} }

\maketitle

\begin{abstract}

A significant challenge in racing-related research is the lack of publicly available datasets containing raw images with corresponding annotations for the downstream task. In this paper, we introduce RoRaTrack, a novel dataset that contains annotated multi-camera image data from racing scenarios for track detection. The data is collected on a Dallara AV-21 at a racing circuit in Indiana, in collaboration with the Indy Autonomous Challenge (IAC). RoRaTrack addresses common problems such as blurriness due to high speed, color inversion from the camera, and absence of lane markings on the track. Consequently, we propose RaceGAN, a baseline model based on a Generative Adversarial Network (GAN) that effectively addresses these challenges. The proposed model demonstrates superior performance compared to current state-of-the-art machine learning models in track detection. The dataset and code for this work are available at https://github.com/ghosh64/RaceGAN.

\begin{IEEEkeywords} Machine Learning, Deep Learning, Lane Detection, Track Detection, Autonomous Racing
\end{IEEEkeywords}

\end{abstract}
\IEEEpeerreviewmaketitle

\section{Introduction}

Modern vehicles are increasingly equipped with a range of computer vision technologies to assist drivers and improve road safety. A critical application of these technologies, particularly for autonomous and self-driving vehicles, is lane detection, which ensures that vehicles remain within designated lanes \cite{nuscenes}. Lane detection systems not only help maintain proper lane alignment, but also provide visual cues to drivers about lane boundaries.

Similarly, autonomous technologies are being integrated into race cars, giving rise to the emerging field of autonomous racing. In this domain, vehicles operate entirely without human intervention, relying solely on artificial intelligence and computer vision algorithms \cite{ghosh2024weighted}. In such high-stakes environments, precise and timely execution of tasks such as track detection is crucial, as there is no human driver to correct potential errors.

Autonomous racecars are equipped with a range of sensors that deliver real-time data, enabling the vehicle to understand its environment effectively. Some common sensors include LiDAR, RADAR, GPS/GNSS, and cameras, among others \cite{racecar}. While LiDAR is effective for track detection on race tracks with bounding walls, it struggles on road courses that lack such clear boundaries \cite{ghosh2024weighted}.  Road courses, often surrounded by grass or gravel, mimic real-world roads and present unique challenges for LiDAR-based systems. In addition, all sensors have inherent error margins and could malfunction, potentially disrupting navigation. These limitations highlight the need for reliable camera-based methods to detect track boundaries, which serve as affordable alternatives and augmentations of existing systems.

\begin{table*}[!htb]
    \centering
    \caption{Comparison of our RoRaTrack dataset and existing traffic datasets for lane and track detection.}
    \begin{tabular}{
        >{\columncolor[HTML]{CFE2F3}}c  
        c
        c
        c
        c
        c
        c
    }
        \toprule
        \rowcolor{gray!30}  
        \textbf{Datasets} & \textbf{Year} & \textbf{Type} & \textbf{Camera Angle} & \textbf{Annotations} & \textbf{Dataset Type} &\textbf{Data Type}\\
        \midrule
        GTA5\cite{gta5}  & 2016 & Traffic & Front Dash & Masks & Lane Detection &Real\\
        SYNTHIA\cite{synthia} & 2016 & Traffic & Multi & Masks & Lane Detection &Synthetic\\
        TuSimple\cite{TuSimple} & 2017 & Traffic & Front Dash & Bounding Lines & Lane Detection &Real\\
        CULane\cite{CULane} & 2018 & Traffic & Front Dash & Bounding Lines & Lane Detection &Real\\
        LLAMAS\cite{llamas} & 2019 & Traffic & Front Dash & Bounding Lines & Lane Detection &Real\\
        nuScenes\cite{nuscenes} & 2020 & Traffic & Front Dash & Semantic Segmentation & Lane Detection &Real\\
        BDD100K\cite{bdd100k} & 2020 & Traffic & Front Dash & Masks, Bounding Lines & Lane Detection &Real\\
        a2d2\cite{a2d2} & 2020 & Traffic & Front Dash & Semantic Segmentation & Lane Detection &Real\\
        VIL-100\cite{vil100} & 2021 & Traffic & Front Dash & Bounding Lines & Lane Detection &Real\\
        RACECAR \cite{racecar} & 2023 & Racing & Multi & None & Track Detection &Real\\
        WeBACNN \cite{ghosh2024weighted} & 2024 & Racing & Multi & Masks & Track Detection &Synthetic\\
        \textbf{RoRaTrack} (Ours) & 2024 & Racing & Multi & Masks & Track Detection &Real\\
        \rowcolor{gray!10}
        \bottomrule
    \end{tabular}
    
    \label{tab:dataset_comparisons}
\end{table*}



Despite growing interest in autonomous racing, there is a notable lack of datasets specifically designed for racing environments. Existing traffic datasets, while useful for urban scenarios, fail to address the unique challenges of racing, such as high speeds, blurred images, and the absence of clear lane markings. These factors make it difficult for models trained on traffic data to generalize effectively to racing conditions.

In this paper, we focus on track detection as a key task and address these challenges through the following contributions:

\begin{enumerate}
    \item We introduce the \textbf{Road Racing Track Dataset (RoRaTrack), the first open source dataset specifically designed for autonomous racing on road courses.} This dataset includes image data paired with instance-level annotations in the form of segmentation masks, collected using cameras mounted at various angles on an autonomous race car.  RoRaTrack captures common racing challenges, including single-lane tracks, lack of lane markings, high-speed data, and image distortion.
    \item We propose \textbf{RaceGAN}, a GAN-based method for track detection tailored to the unique challenges of racing environments. Through a comprehensive evaluation of eight state-of-the-art track detection methods, we demonstrate that RaceGAN significantly outperforms existing approaches, setting a new benchmark for track detection in autonomous racing. 
\end{enumerate}



\section{Literature Review}

\subsection{Track Detection Methods }
In this section, we focus on existing work on track detection methods for autonomous racing vehicles, which is a specialized application of segmentation in road scenes. We first review state-of-the-art segmentation methods, then segmentation in road scenes, and finally, track detection methods. 




\textit{Segmentation for General Tasks:} YOLOv8\cite{yolov8} excels in both object detection and segmentation, combining CSPDarknet for feature extraction, PANet for feature aggregation, and a U-Net inspired decoder. With cIoU and DiceLoss, it achieves state-of-the-art segmentation. However, its  requirement of a large memory limits its deployment on resource-constrained devices. Hetnet\cite{hetnet} integrates low-level features (e.g., intensity contrast) with high-level contextual information for mirror detection. SAM2-Unet\cite{sam2-unet}, based on the Segment Anything Model (SAM2)\cite{sam2}, uses SAM2 as an encoder for U-Net architectures, showing strong segmentation performance with a Hiera backbone and U-shaped decoder. Reseg\cite{reseg} combines CNN-extracted local features with the long-range dependency modeling of Recurrent Neural Networks (RNNs) using the ReNet\cite{renet} model for segmentation tasks. DeepLabv3\cite{deeplabv3}
, revisits atrous convolutions to manage the filter’s field-of-view and capture multi-scale context, improving segmentation across varying object sizes.


\textit{Segmentation for Road Scenes:} WeBACNN\cite{ghosh2024weighted} employs a weighted branch aggregation strategy to combine local and global features through context-aware aggregation. It frames track detection as a segmentation task. SegNet\cite{segnet} is designed for efficient pixel-wise semantic segmentation in road scenes. It consists of an encoder and decoder network, with the encoder architecture based on the 13 convolutional layers of the VGG16 network\cite{vgg16}, followed by a pixel-wise classification layer. Ultrafast\cite{ultrafast} is a lane detection method optimized for speed. Instead of pixel-wise segmentation, it identifies lane locations, making it faster compared to deep segmentation models. It also uses global features, providing a larger receptive field compared to traditional segmentation approaches.

\textit{Lane Detection Methods:} Automatic Lane Detection Methods in the literature can be broadly categorized into three types: parameter-based, anchor-based, and segmentation-based approaches. Parameter-based methods typically involve techniques such as polynomial curve fitting, as seen in \cite{Feng_2022_CVPR, poly}, where the lane boundaries are modeled using mathematical equations. Anchor-based methods include models like Line-CNN\cite{linecnn} and LaneAtt\cite{laneatt}, both of which rely on predefined lane line suggestions. Line-CNN utilizes a Convolutional Neural Network (CNN), while LaneAtt employs an attention mechanism to refine lane predictions. As mentioned above, racing scenarios may lack clearly defined lane boundaries, which is why these methods are ineffective in our case. Segmentation-based methods treat lane detection as a pixel-wise segmentation problem. For example, \cite{Zheng_2022_CVPR} integrates both high- and low-level features by incorporating the global context and refining it with low-level details. Similarly, \cite{vil100} introduces a multi-level memory aggregation network, which enhances feature representation across different scales. \cite{Wang_2022_CVPR} proposes the Global Association Network (GANet), where lane key points are directly regressed to their starting positions, and the Lane-aware Feature Aggregator improves local predictions by incorporating global context. 

\begin{figure*}[!htb]
\begin{tabular}{
        >{\columncolor[HTML]{FFFFFF}}c  
        c
        c
        c
        c
        c}

 \includegraphics[width=1.02in]{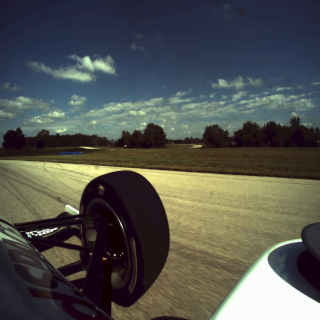}
 &\includegraphics[width=1.02in]{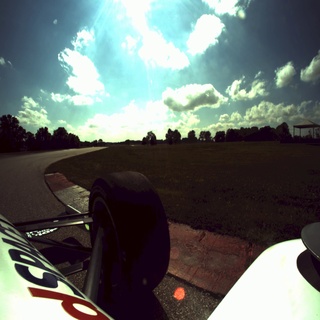}
 &\includegraphics[width=1.02in]{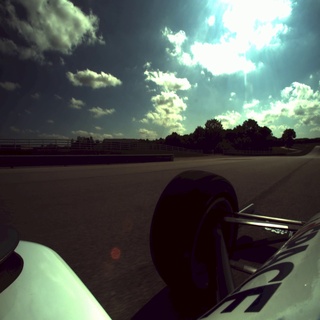}
 &\includegraphics[width=1.02in]{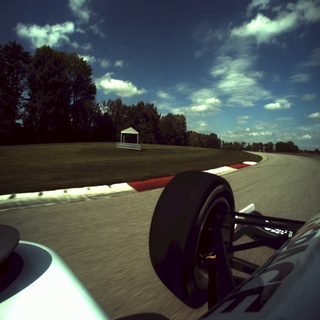}
 &\includegraphics[width=1.02in]{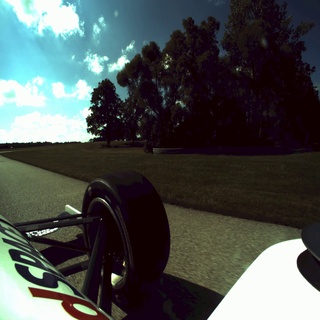}
 &\includegraphics[width=1.02in]{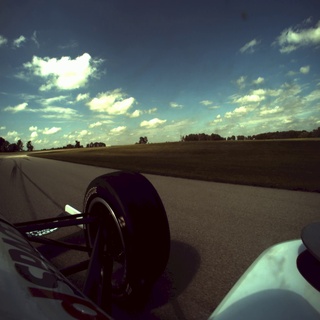} \\

 Normal &Dazzle Light & \parbox[t]{80pt}{Color Imbalance\\ - Green Hue} &Curved Road & \parbox[t]{80pt}{Color Imbalance\\ - Underexposed} &Blurry \\
\end{tabular}
\caption{Images depicting normal and challenging road scenarios—such as dazzle light, color imbalance, curved roads, and blurriness. \vspace{0pt}}
\label{fig:dataset_composition}
\end{figure*}

\subsection{Existing Datasets}


Table \ref{tab:dataset_comparisons} shows a detailed comparison of existing lane detection datasets. In the last row, we also provide a comparison with the proposed RoRaTrack dataset. 

Most of the existing datasets are annotated specifically for lane detection, and only two are based on racing datasets. They often provide annotations of the bounding lines, as seen in TuSimple \cite{TuSimple}, CULane \cite{CULane}, LLAMAS \cite{llamas}, and VIL-100 \cite{vil100}, which makes them unsuitable for track detection algorithm development. While semantic segmentation or masks are available for lane identification for some datasets such as GTA5 \cite{gta5}, nuScenes \cite{nuscenes}, BDD100k \cite{bdd100k}, and a2d2 \cite{a2d2}, the characteristics of these traffic datasets differ significantly from those of racing data, with all of them providing data from only the front dash camera. While SYNTHIA \cite{synthia} lane detection dataset provides multi camera data, the data is synthetically generated, and not real.

While RACECAR \cite{racecar} offers a racing dataset, it is stored in rosbags, which are not directly usable and lack annotations. WeBACNN \cite{ghosh2024weighted} provides a racing dataset that includes images exhibiting the common challenges of racing data. However, this dataset was curated using animated racing videos, which may limit the transferability of model learning to real-world datasets. 

As the table indicates, there is currently no dataset that meets all the following criteria:
\begin{itemize}
    \item Built specifically for track detection in racing environments.
    \item Annotated with segmentation masks that clearly separate the road from the background.
    \item Has a multi-camera feed.
    \item Provides real-life data (non-synthetic).
\end{itemize}
In this paper, we address this vital gap in existing datasets, by providing a track detection dataset that includes two camera angles, with data collected from real-life high-speed racing scenarios, and provides lane segmentation masks as annotations.

\section{RoRaTrack Dataset}

In this section, we provide a detailed description of the construction, characteristics, and annotations included in the RoRaTrack dataset.

\subsection{Data Collection and Split}
This dataset was collected using the Dallara AV-21 race car on Putnam Park Road Course, Putnam County, Indiana. The Dallara AV-21 has a Carbon Chassis with a race weight of 640-649 Kg, 335kW/450 Horsepower, Ricardo 6-speed semi-automatic gearbox, rear-wheel drive. The sensors onboard include 6 mono cameras (2 front, 2 stereo, 2 rear), 4 Radars, 3 LiDARs, and a TRK GPS. The computing platform is an Intel Xeon CPU equipped with an NVIDIA Quadro RTX 8000 GPU. 

Data collected on a road course is considerably different from that collected on a motor speedway. The track layout in a motor speedway is usually oval and generally at the same elevation throughout the track, often containing lane markings for multi-vehicle racing. On the other hand, road courses like Putnam Park have multiple winding turns with some elevation differences over the course. 


The data was collected from cameras on one run of the Dallara AV-21 vehicle on the track that took approximately 12 minutes. The vehicle is equipped with 4 video cameras \textemdash front right, front left, rear right, and rear left. The data in the RoRaTrack dataset has been collected from the front left and front right cameras. The sampling rate of the camera is 30 frames per second (fps). We down-sample the 30 fps videos to 1 fps for both left and right videos to yield 607 front left images and 791 front right images. The total number of images in the dataset is 1398 which is divided into a training and testing split of 80-20.

We utilize perception-based track detection to identify and map the unobstructed area in front of the vehicle, enabling safe and efficient navigation. This becomes particularly crucial during curves and turns, where the vehicle must accurately determine the available space that it can occupy and adjust its speed accordingly. For this purpose, we used data from the two front cameras to train our model. 

\begin{table}[t]
    \centering
    \caption{Dataset composition showing the distribution of images across various categories. Apart from \textit{Normal}, the other categories are not mutually exclusive, meaning a single image may belong to multiple categories.  \vspace{0pt}}
    \begin{tabular}{
        >{\columncolor[HTML]{CFE2F3}}c  
        c
        c
    }
        \toprule
        \rowcolor{gray!30}  
        \textbf{Categories} & \textbf{No. of Images}  \\
        \midrule
        Normal & 146  \\
        Curved Roads & 378  \\
        Color Imbalance &350  \\
        Blurry & 557 \\
        Dazzle Light & 745 \\
        \rowcolor{gray!10}
        \hline
        Total & 1398 \\
        \bottomrule
    \end{tabular}
    \label{tab:dataset_composition}
\end{table}

\begin{figure*}[!htb]
    \centering
    \includegraphics[width=\textwidth]{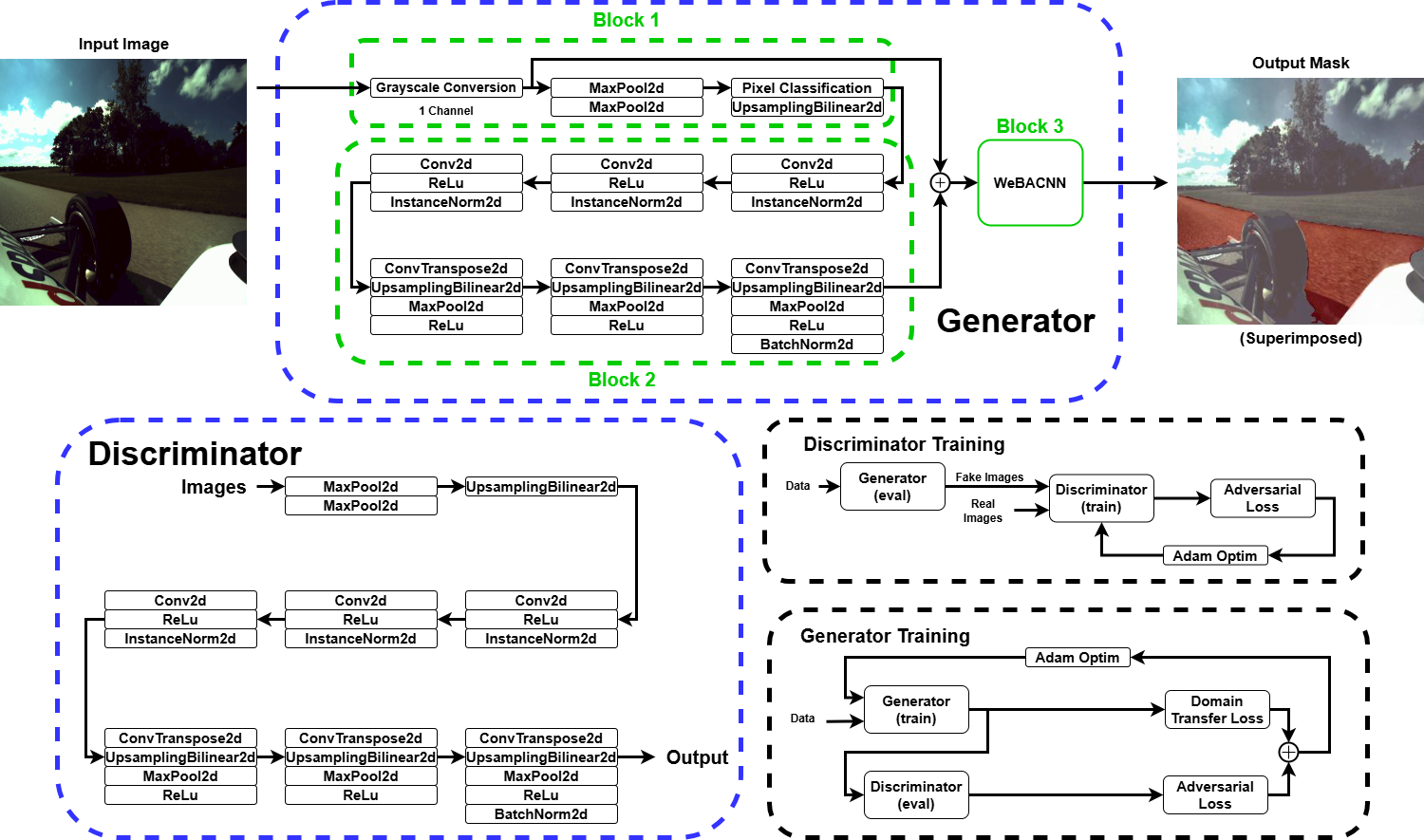}
    \caption{The architecture of the proposed RaceGAN model is depicted, highlighting the detailed structure of both the generator and discriminator blocks. Additionally, the training flow for the discriminator and the generator is illustrated separately. \vspace{-10pt}}
    \label{fig:gan_architecture}
\end{figure*}

\begin{figure*}[!htb]
    \centering\includegraphics[width=\textwidth]{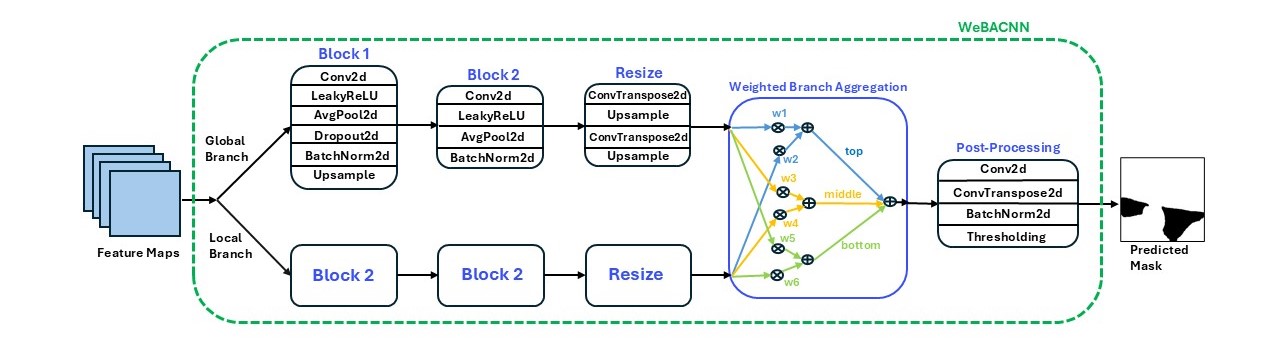}
    \caption{Architecture of the WeBACNN block \cite{ghosh2024weighted}. \vspace{0pt}}
    \label{fig:wbacnn}
\end{figure*}

\subsection{Dataset Characteristics}

In Table \ref{tab:dataset_composition}, the distribution of the data is presented in the different racing scenarios. Examples of these diverse scenarios are shown in Figure \ref{fig:dataset_composition}. As a purpose-built dataset for autonomous racing on road courses, RaceGAN incorporates the unique characteristics present in this challenging environment. Frames for videos collected at high speed and in different weather conditions suffer from blurriness, color imbalance (over- or under-exposed), and dazzle light. Images also have saturation-related artifacts, such as a green hue. Due to these factors, only 146 images out of 1398 images can be classified as relatively normal (straight road, no color imbalances or blurriness). The data was collected on a road course that are known for their windy turns, and 378 images captured curved roads. 

Traffic datasets typically collect data from a single front-facing camera, which offers a clear view of road and lane markings. In contrast, our dataset \textemdash collected from a race car \textemdash includes views from both the right and left, providing a perspective more suited for track detection in a racing environment. Since our dataset is intended to perform track detection for the explicit purpose of racing, it does not contain any lane markings or lane boundaries. Additionally, the chassis of the vehicle occludes the lane in all images, while also capturing a significant portion of the background. 

The RoRaTrack dataset offers a unique and challenging environment for autonomous racing research, with diverse and realistic scenarios, multiple camera views, and complex visual conditions, making it an ideal resource for developing and testing robust track detection algorithms.


\subsection{Annotation}

We employed a two-stage annotation approach for our dataset. Initially, 20\% of the RoRaTrack dataset was manually annotated with segmentation masks. A pre-trained YOLOv8n-seg model, initially trained on the COCO dataset, was subsequently fine-tuned using the annotated data. In the second stage, the fine-tuned YOLOv8n-seg model was utilized to generate segmentation masks for the remaining 80\% of the dataset, which served as our ground truth annotations. To further validate the accuracy of the annotations, we performed random manual inspections on the segmentation masks produced by the YOLOv8n-seg model. 

\section{Proposed Method}

Our proposed RaceGAN model, illustrated in Figure \ref{fig:gan_architecture}, leverages a Deep Convolutional Generative Adversarial Network (DCGAN) architecture, which consists of both deep convolutional generator and discriminator networks. This model integrates key elements of CycleGAN \cite{cyclegan} and Wasserstein GAN \cite{wgan}, combining their strengths to improve performance. Specifically, the design choices are carefully tailored to optimize track recognition, enabling more effective learning and generation of track patterns.

The Generative Adversarial Network (GAN), utilized in RaceGAN, is a type of deep learning network that generates new data samples based on a target training dataset. GANs are made up of two neural networks: a generator and a discriminator. The two networks are trained simultaneously as adversaries. The goal of the generator network is to generate realistic synthetic images that make it difficult for the discriminator to distinguish between real and fake images. In parallel, the discriminator is trained to distinguish real data from fake data. This adversarial process leads the generator to improve over time and ultimately generate realistic data that closely resembles the training dataset.

\subsection{Generator}
The generator architecture is based on the WeBACNN model (Figure \ref{fig:wbacnn}) which has shown strong performance in track detection tasks \cite{ghosh2024weighted}. We improve the model by augmenting it with additional computational blocks designed for deep feature extraction, as well as residual connections to enhance the generation of realistic images. Figure \ref{fig:gan_architecture} shows the detailed construction of these two blocks.

The first block of the generator is tasked with making an initial prediction regarding the pixels, determining whether they belong to a region containing lane. This process requires an understanding of the composition of the image. Typically, in an image, the road surface exhibits a relatively uniform color, while non-lane regions are characterized by multiple colors. This block is designed to classify regions with diverse pixel values as non-lane areas and regions with more uniform (monochromatic) pixel values as lane regions. To achieve this, the image is first converted to grayscale. Then, two downsampling layers, each using kernels of different sizes, are applied to capture both global statistics and local details in a region. To further refine the pixel classification, a custom classification algorithm is introduced, which enhances the distinction between dark pixels (likely to be part of the lane) and bright pixels (typically representing the background).

The second block focuses on deep feature extraction from the initial guesses. This block consists of consecutive convolution layers with different kernel sizes to capture different levels of detail. The smaller kernels facilitate the extraction of fine-grained details within localized regions of the image, whereas the larger kernels provide more contextual information about the region. Each convolutional block also contains normalization layers for better generalizability and stable training. 


The result is then passed into the WeBACNN backbone, which makes up the last block and generates the final prediction. This model contains two branches for two different layers of detail. During aggregation each branch is weighted according to the section in which it is predicting the mask. For sections that require more fine details, the branch containing more localized features is assigned a greater weight to ensure a smooth reconstruction, whereas for sections that require more global context, the branch containing higher-level features is assigned a greater weight. Together, they utilize regional weights in different parts of the image to give a smoother prediction.

\subsection{Discriminator}

The discriminator is composed of convolutional layers, max-pooling layers, normalization, and transpose convolutions, which collectively enable deep feature extraction, similar to the second block of the generator architecture. The discriminator's role is to classify each pixel as real or fake, with the average classification of all pixels being used to determine the authenticity of the image. Instead of explicitly classifying the entire image, the discriminator operates more like a 'critic,' evaluating each pixel independently. This approach leads to more refined and accurate outputs.


\begin{figure*}[t]
    \centering
    \subfloat[\label{fig:original_image}]{%
       \includegraphics[width=0.23\linewidth]{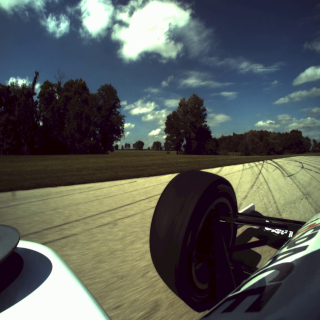}}
    \hspace{12pt}
    \subfloat[\label{fig:lane_mask}]{%
        \includegraphics[width=0.23\linewidth]{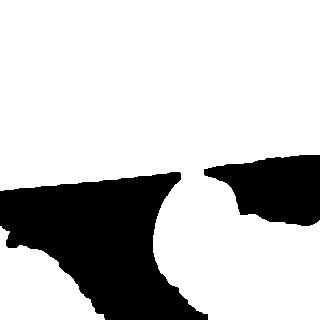}}
    \hspace{12pt}
    \subfloat[\label{fig:superimposed_image}]{%
        \includegraphics[width=0.23\linewidth]{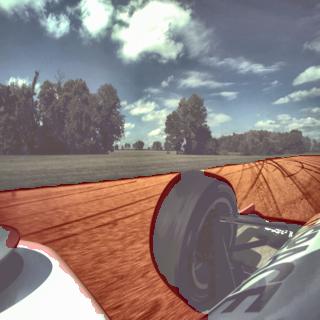}}
        
    \caption{Representative images illustrating the track prediction process: (a) The original input image, (b) The predicted lane mask, and (c) The superimposed image displaying the predicted lane (highlighted in red) overlaid on the original image. \vspace{-5pt}}
    \label{fig:superimposed_key}
\end{figure*}

\subsection{GAN Training}
Diverging from traditional GAN architectures, RaceGAN takes an image of the track as input, rather than a random noise vector. This design choice enables the generator to produce domain-specific images. To reinforce this domain specificity, we introduce a domain transfer loss and an adversarial loss, as formulated in Equation \ref{eq:ganloss}, to guide the GAN training process.

\begin{equation}
    \mathcal{L}_{\text{total}} = \mathcal{L}_{\text{adv}}(G, D) + \lambda \mathcal{L}_{\text{domain}}(G, F)
    \label{eq:ganloss}
\end{equation}

\begin{itemize}
    \item \( \mathcal{L}_{\text{adv}}(G, D) \): The adversarial loss, where \( G \) is the generator and \( D \) is the discriminator. This term helps the generator create realistic outputs by trying to mislead the discriminator. This loss is used to train the discriminator. 
    \item \( \lambda \): A hyperparameter that controls the balance between the adversarial loss and the domain transfer loss.
    \item \( \mathcal{L}_{\text{domain}}(G, F) \): The domain transfer loss where \( G \) is the generator.
\end{itemize}

The domain transfer loss function assesses the alignment of generated data with the target domain, using an L2 norm to measure the difference between generated and real target domain data. During generator training, the overall loss is a weighted combination of the adversarial loss, which incentivizes the generator to deceive the discriminator, and the domain transfer loss, which enforces the desired domain adaptation. By fixing the discriminator, the generator learns to produce realistic and domain-specific images simultaneously. In contrast, during discriminator training, the generator is kept constant, and the discriminator learns only through adversarial loss. Throughout training, the accuracy of the discriminator remains near 50\%, indicating its inability to reliably distinguish real from fake images, characteristic of GANs in equilibrium.

\subsection{Post Processing}
To improve output quality, a postprocessing step is introduced to reduce noise in the generated results. This process involves analyzing each pixel and its neighbors, evaluating the connectivity of the pixels to remove small scattered groups, and then applying two morphological operations to the remaining pixels.

To enhance the quality of the output, a post-processing step is implemented to minimize noise in the generated results. This step involves examining each pixel and its surrounding neighbors, assessing pixel connectivity to eliminate small scattered groups, and then applying two morphological operations to the remaining pixels.

The first operation, dilation, expands each pixel group, bridges the gaps between adjacent clusters. It works by placing a structuring element over each foreground pixel and setting the corresponding output pixels within the structuring element to foreground. Mathematically, this can be represented as Equation \ref{eq:dilation}.

\begin{equation}
    A \oplus B = \{ z \mid (B)_z \cap A \neq \emptyset \}
    \label{eq:dilation}
\end{equation}

\begin{itemize}
    \item \( A \): The binary image (or foreground pixels) being dilated.
    \item \( B \): The structuring element, a small binary mask used to expand the foreground.
    \item \( (B)_z \): The structuring element \( B \) translated (shifted) to position \( z \).
    \item \( z \): Position within the image to which the structuring element B is shifted.
\end{itemize}

The second operation, erosion, shrinks the size of connected pixel groups, smoothing and refining their boundaries. This is accomplished by positioning a structuring element over each foreground pixel; if the element does not fit fully within the foreground, the pixel is set to background. Mathematically, this is represented as Equation \ref{eq:erosion}.

\begin{equation}
    A \ominus B = \{ z \mid (B)_z \subseteq A \}
    \label{eq:erosion}
\end{equation}
\begin{itemize}
    \item \( A \): Input image 
    \item \( B \): Structuring element (shape/kernel used for erosion)
    \item \( (B)_z \): Translation of \( B \) at position \( z \)
   \item The pixel in the output image is part of the foreground if the structuring element \( B \) fits completely within the image at position \( z \).
\end{itemize}




\section{Results}\label{results}

We provide the results of qualitative and quantitative analysis of our model's performance compared against eight competing methods for track detection. To the best of our knowledge, there are currently no track detection models specifically trained on road racing data. Therefore, a direct comparison with specialized track detection methods for road racing data is not possible. However, WeBACNN has demonstrated success in track detection tasks for racing circuits. Additionally, several state-of-the-art methods for traffic lane detection are readily available. To establish benchmark results, we select seven traffic lane detection methods, and one track detection method (WeBACNN), and train and test them on the RoRaTrack dataset. Subsequently, we compare the results with those obtained using RaceGAN. Some methods were augmented with a post-processing thresholding method to enhance the accuracy of the detections.

All experiments were conducted on a computer with the following specifications: Intel Xeon Gold 6142 CPU with 64 cores, 256GB of RAM, and a Nvidia Tesla P100 GPU with 16GB of memory.



\begin{table*}[t]
    \centering
    \caption{A detailed comparison of track detection performance for our proposed method and other state-of-the-art techniques. Methods annotated with $^{\ast}$ indicate the use of postprocessing steps to enhance their results. The top-performing metrics across all methods are emphasized in bold for clarity. \vspace{0pt}}
    \begin{tabular}{
        >{\columncolor[HTML]{CFE2F3}}c  
        c
        c
        c
        c
        c
        c
    }
    \toprule
        \rowcolor{gray! 30}  
        \textbf{Method} & \textbf{mIoU} & \textbf{Accuracy} & \textbf{Precision} & \textbf{Recall} & \textbf{F1 Score} & \textbf{Specificity} \\
        \midrule
        WeBACNN$^{\ast}$ & 0.8268 & 0.9480 & 0.9210 & 0.7528 & 0.8168 & 0.9895  \\
        Ultrafast & 0.0876 & 0.1753 & 0.1753 &\textbf{1} & 0.2961 & 4.053$ \times 10^{-6}$ \\
        Segnet & 0.7200 & 0.9171 & \textbf{0.9687} & 0.5416 & 0.6795 & 0.965 \\
        Reseg$^{\ast}$ & 0.7890 & 0.9360 & 0.9032 & 0.6976 & 0.7711 & 0.9864 \\
        Polylanenet$^{\ast}$ & 0.4142 & 0.7809 & 0.1750 & 0.0673 & 0.0956 & 0.9325 \\
        Hetnet$^{\ast}$ & 0.8482 & 0.9547 & 0.9406 & 0.7842 & 0.8416 & \textbf{0.9908} \\
        Deeplabv3$^{\ast}$ & 0.6554 & 0.8545 & 0.5698 & 0.5950 & 0.5769 & 0.9047 \\
        Sam2-unet & 0.3340 & 0.6454 & 0.0494 & 0.0547 & 0.0509 & 0.77 \\
        \rowcolor{gray!10}
        \textbf{RaceGAN} & \textbf{0.8691} & \textbf{0.9580} & 0.8552 & 0.9059 & \textbf{0.8738} & 0.9682 \\
        \bottomrule
    \end{tabular}
    \vspace{-15pt}
    \label{tab:comparison_methods}
\end{table*}

\subsection{Evaluation Metrics}

For the quantitative analysis, we use the following metrics to evaluate the models\textemdash Mean Intersection over Union (mIoU), Accuracy, Precision, Recall, F1 score and Specificity. For each metric, we assume that each image pixel can be classified into two classes, $lane$ or $background$. 

We define a few useful terms that are used in the computation of each metric:

\begin{enumerate}
    \item \textit{True Positives (TP)} Number of pixels correctly classified as the positive class. This corresponds to the number of pixels correctly classified as lane.
    
    \item \textit{True Negatives (TN)} Number of pixels correctly classified as the negative class. This corresponds to the number of pixels correctly classified as background.
    
    \item \textit{False Positives (FP)} Number of pixels wrongly classified as the positive class. This number represents the number of background pixels classified as lane pixels.
    
    \item \textit{False Negatives (FN)} Number of pixels wrongly classified as the negative class. This number represents the number of lane pixels classified as background pixels.
\end{enumerate}

Next, we go into the definition and computation of each metric used to evaluate the models.

\begin{enumerate}
    \item \textit{mIoU:} We define the pixel-wise IoU as follows:
\vspace{-2pt}
\begin{equation}
\label{eq:iou}
IoU(pred, label)=\frac{track\_pixels(pred \cap label)}{track\_pixels(pred \cup label)}
\end{equation}

Where $track\_pixels$ gives us the number of pixels that meet the required condition. In this case, we compute the mIoU over two classes \textemdash lane and background. The intersection represents the number of pixels that are marked as lane when computing IoU for lane and background when computing IoU for background. The union represents the total number of pixels that are marked as the class for which we are computing the IoU. Both IoUs are averaged to give us the final mIoU values.

\begin{figure*}[!htb]
\setlength\tabcolsep{2pt}
\begin{tabular}{>{\raggedleft}p{45pt}
                cccccccc}
 \rotatebox[origin=c]{270}{\hspace{-42pt} \parbox[t][][b]{80pt}{\hspace{-0pt} Normal}}
 &\includegraphics[width=0.75in]{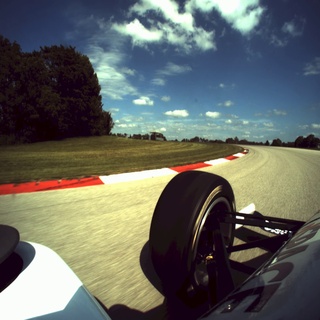}& \includegraphics[width=0.75in]{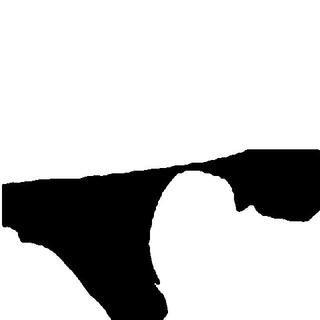}&
 \includegraphics[width=0.75in]{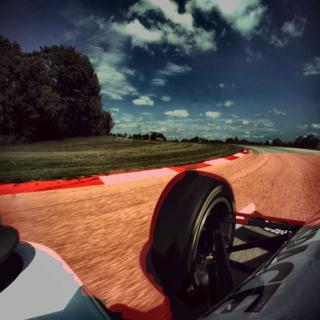}&
 \includegraphics[width=0.75in]{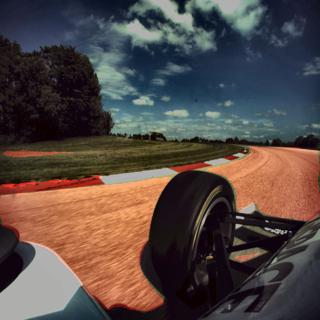}&
 \includegraphics[width=0.75in]{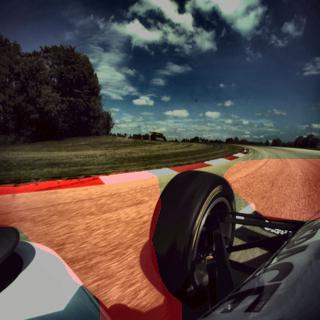}&
 \includegraphics[width=0.75in]{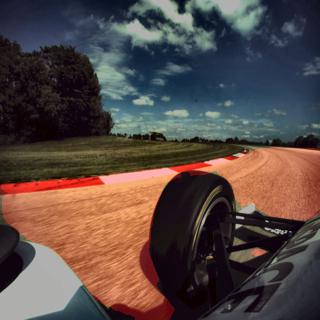}&
 \includegraphics[width=0.75in]{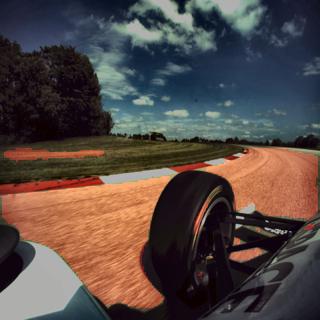}&
 \includegraphics[width=0.75in]{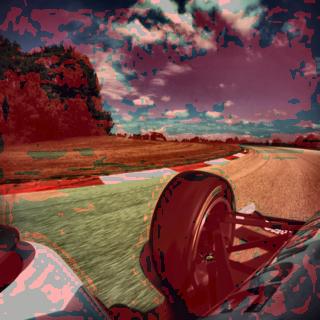} 
 \vspace{-21pt} \\

 \rotatebox[origin=c]{270}{\hspace{-35pt} \parbox[t]{80pt}{Dazzle\\ Light}}
 &\includegraphics[width=0.75in]{imgs/input_images_resized/front_right_frame_0350.jpg}& \includegraphics[width=0.75in]{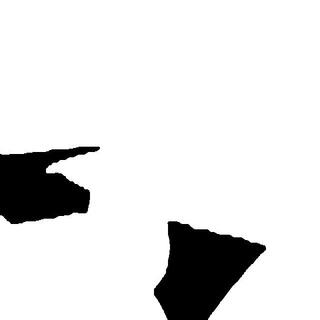}& \includegraphics[width=0.75in]{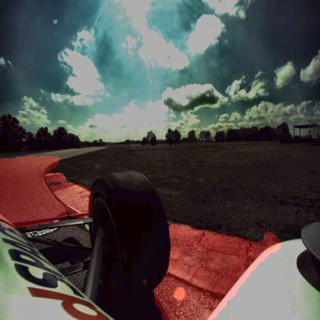}&
 \includegraphics[width=0.75in]{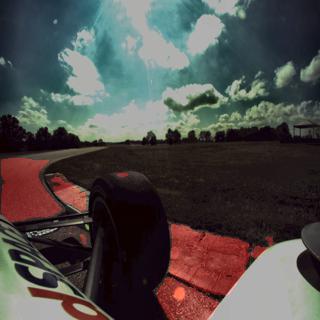}&
 \includegraphics[width=0.75in]{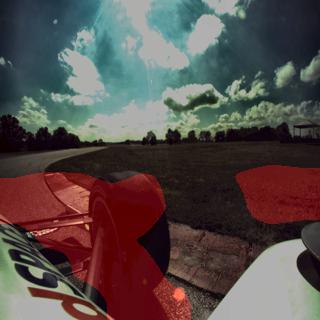}&
 \includegraphics[width=0.75in]{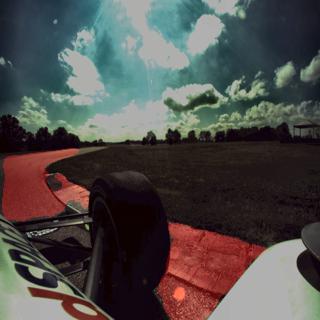}&
 \includegraphics[width=0.75in]{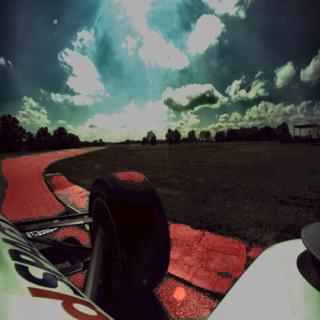}&
 \includegraphics[width=0.75in]{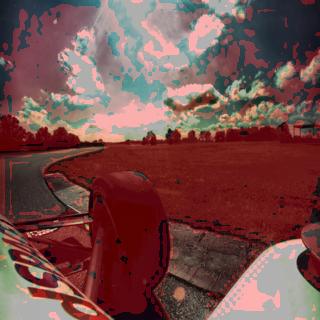}
 \vspace{-24pt} \\
 
 \rotatebox[origin=c]{270}{\hspace{-45pt} \parbox[t]{80pt}{Color\\Imbalance\\ (Green Hue)}} &\includegraphics[width=0.75in]{imgs/input_images_resized/front_left_frame_0411.jpg}& \includegraphics[width=0.75in]{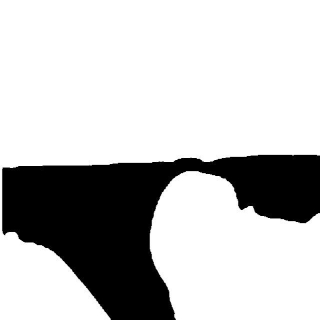}& \includegraphics[width=0.75in]{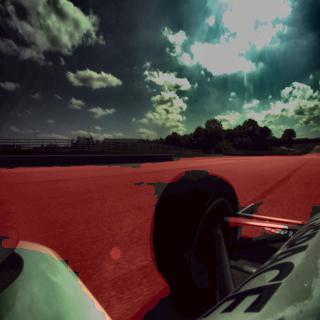}&
 \includegraphics[width=0.75in]{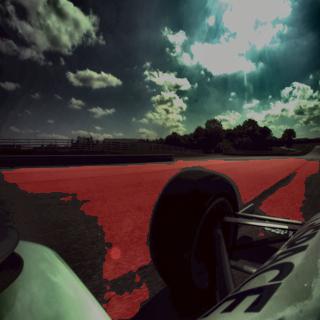}&
 \includegraphics[width=0.75in]{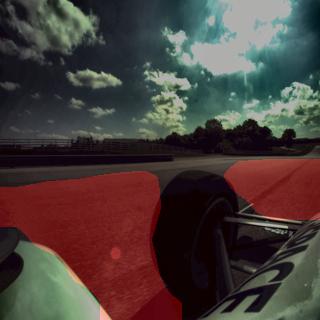}&
 \includegraphics[width=0.75in]{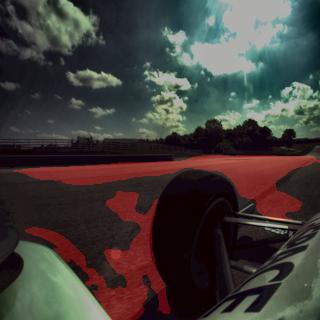}&
 \includegraphics[width=0.75in]{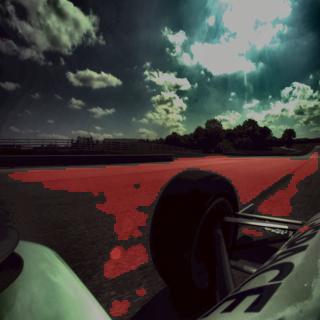}&
 \includegraphics[width=0.75in]{imgs/sam2_unet/result_front_left_frame_0532.jpg}
 \vspace{-26pt} \\
 
 \rotatebox[origin=c]{270}{\hspace{-30pt} \parbox[t]{80pt}{Curved\\ Road}}&\includegraphics[width=0.75in]{imgs/input_images_resized/front_left_frame_0536.jpg}& \includegraphics[width=0.75in]{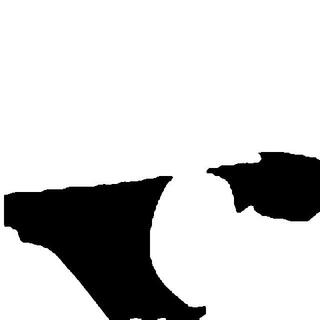}& \includegraphics[width=0.75in]{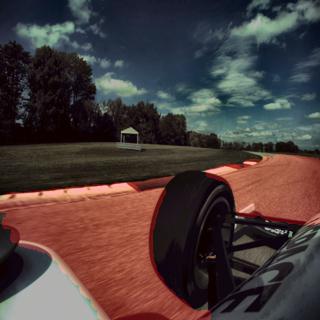}&
 \includegraphics[width=0.75in]{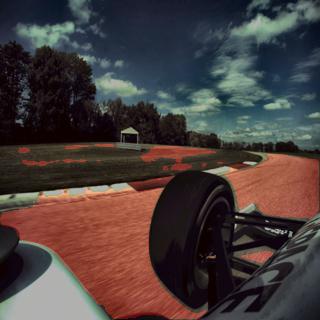}&
 \includegraphics[width=0.75in]{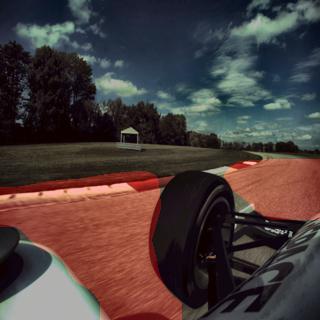}&
 \includegraphics[width=0.75in]{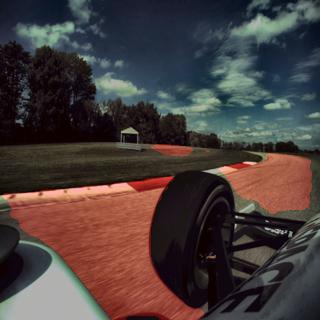}&
 \includegraphics[width=0.75in]{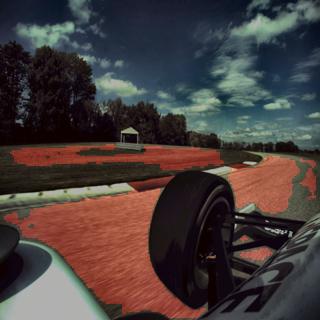}&
 \includegraphics[width=0.75in]{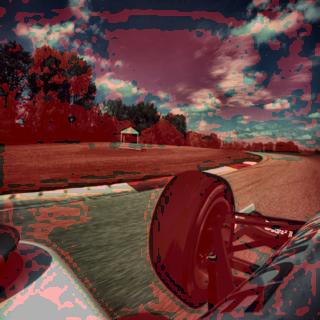}
 \vspace{-26pt} \\
 
 
 \rotatebox[origin=c]{270}{\hspace{-55pt} \parbox[t]{80pt}{\raggedright Color \\Imbalance\\(Underexpo\\ -sed)}} &\includegraphics[width=0.75in]{imgs/input_images_resized/front_right_frame_0012.jpg}& \includegraphics[width=0.75in]{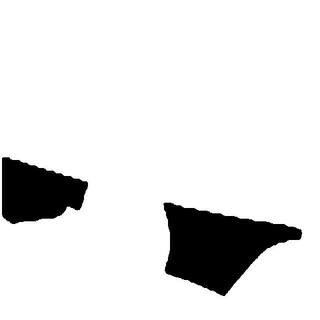}& \includegraphics[width=0.75in]{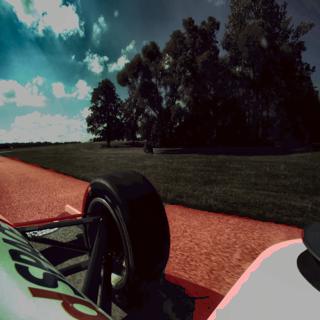}&
 \includegraphics[width=0.75in]{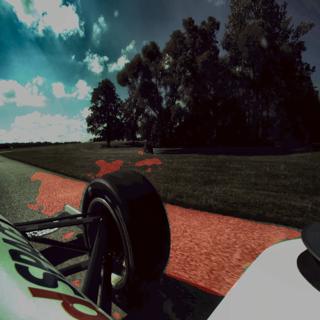}&
 \includegraphics[width=0.75in]{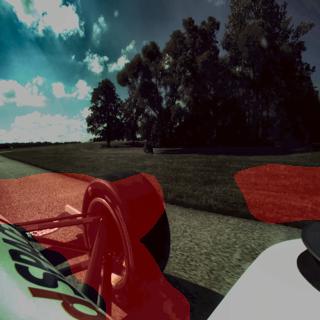}&
 \includegraphics[width=0.75in]{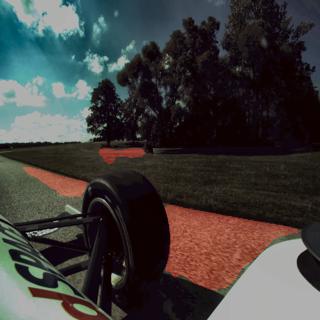}&
 \includegraphics[width=0.75in]{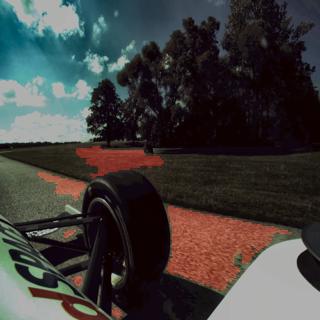}&
 \includegraphics[width=0.75in]{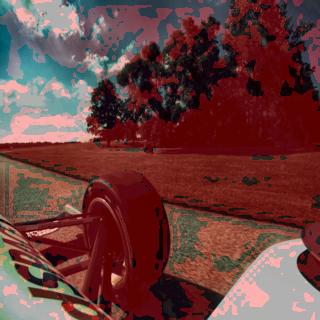}
 \vspace{-26pt} \\
 
 \rotatebox[origin=c]{270}{\hspace{-22pt} \parbox[t]{80pt}{Blurry}} &\includegraphics[width=0.75in]{imgs/input_images_resized/front_right_frame_0112.jpg}& \includegraphics[width=0.75in]{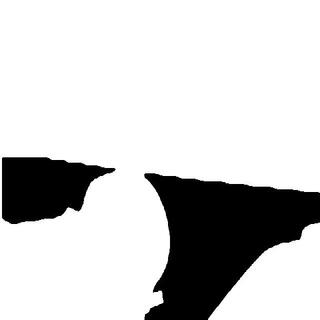}& \includegraphics[width=0.75in]{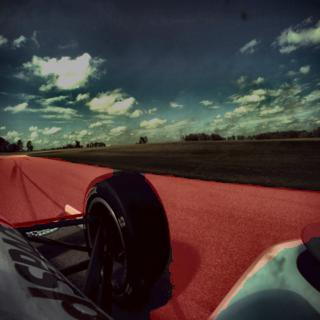}&
 \includegraphics[width=0.75in]{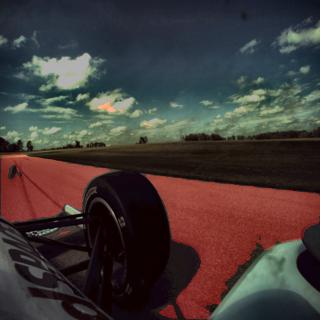}&
\includegraphics[width=0.75in]{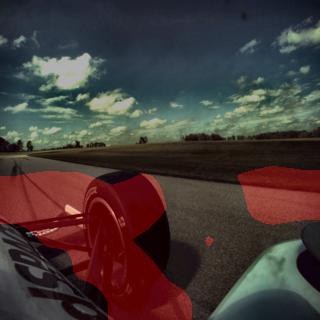}&
 \includegraphics[width=0.75in]{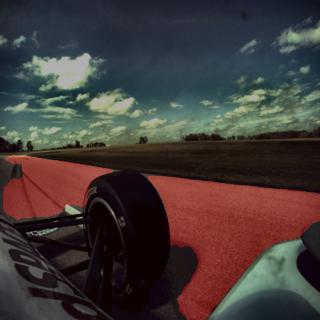}&
 \includegraphics[width=0.75in]{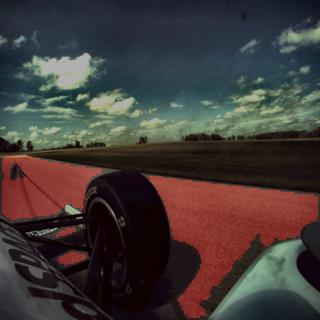}&
 \includegraphics[width=0.75in]{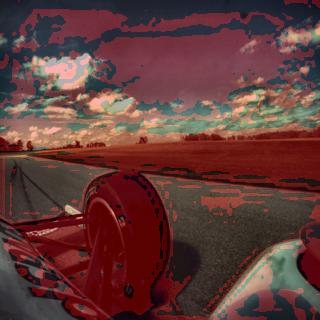}
 \vspace{-26pt} \\

&Input Image &Label & RaceGAN &WeBACNN &Deeplabv3 &HetNet &Reseg &Sam2-Unet\\
\end{tabular}
\caption{ Qualitative comparison of track detection maps produced by the proposed RaceGAN model (3rd column), and state-of-the-art methods (4th to 8th columns), against ground truths (2nd column); with an example of each of the scenarios. \vspace{-10pt}}
\label{fig:qualitative}
\end{figure*}

\item \textit{Accuracy:} Accuracy here refers to the pixel accuracy which is defined as the proportion of correctly classified pixels (for both classes).  Accuracy is computed as 

\begin{equation}
    Accuracy=\frac{TP+TN}{TP+TN+FP+FN}
\end{equation}

As a metric, accuracy may sometimes be misleading, as the area of background is greater than the area of lane in most input images, leading to a higher number of true negatives compared to true positives. Since the accuracy takes into account correctly classified background pixels, this number will be high even if all the pixels are classified as background pixels. To correctly evaluate the model, the accuracy itself is not enough and needs to be supplemented with other metrics such as Recall, F1 score and Specificity.

\item \textit{Precision:} Precision is defined as the proportion of correctly classified positives. 
\begin{center}
    \begin{equation}
        Precision = \frac{TP}{TP+FP}
    \end{equation}
\end{center}
In this case, it refers to the proportion of pixels that were correctly classified as lane among all pixels that were predicted as lane. 

\item \textit{Specificity:} Specificity or True Negative Rate is defined as the proportion of correctly classified negatives. 

\begin{center}
    \begin{equation}
        Specificity=\frac{TN}{TN+FP}
    \end{equation}
\end{center}
In this case, it refers to the proportion of pixels that were correctly classified as background among all pixels that were actually background.

\item \textit{Recall:} Recall is the True Positive Rate which indicates the number of pixels correctly classified as the positive class among the actual number of pixels in the positive class. Recall has been computed as :

\begin{center}
    \begin{equation}
        Recall=\frac{TP}{TP+FN}
    \end{equation}
\end{center}
In our case, it represents the proportion of pixels correctly classified as lane among all the pixels that were actually lane. 

\item \textit{F1 score:} F1 score combines precision and recall into a single metric to give a more balanced evaluation, especially in cases where one of these metrics may be misleading on its own.  F1 score is computed as

\begin{center}
    \begin{equation}
        F1 score=\frac{2 * Precision *Recall}{Precision+Recall}
    \end{equation}
\end{center}
\end{enumerate}

A score of 1 means perfect precision and recall (no false positives or false negatives). A score of 0 means either precision or recall, or both, is 0 (the model is performing poorly).


\subsection{Performance Comparison}

The quantitative performance of our model and the competing methods are presented in Table \ref{tab:comparison_methods}. It can be observed that RaceGAN outperforms competing methods in mIoU, accuracy, and F1 score, with values of 0.8691, 0.9580, and 0.8738, exceeding the best method in each category by 2\%, 0.35\%, and 3.8\%. Among the benchmark methods, the second-best method is Hetnet in terms of mIoU, accuracy, and F1 Score. Hetnet also achieves the highest specificity value of 0.9908. 

In terms of recall, Ultrafast has a perfect score of 1, which means that it correctly identifies all lane pixels as lane. However, it should be noted that Ultrafast also has a low precision value of 0.1753 and a low specificity value of 4.053*1e-6. This implies that Ultrafast classifies most pixels as lane, which would cause a low true negative rate, low precision, yet high recall. As a result, while the recall value is technically the highest, the lack of balance in the other metrics suggests that this method may not truly represent the most effective approach. The second highest recall among competing methods is achieved by the Hetnet method, which, as indicated by the other metrics, demonstrates a more balanced classification, not labeling all pixels as either lane or background. The highest precision is achieved by Segnet, which has a value of 0.9687. However, SegNet has a low recall value, indicating that it misclassifies many lane pixels as background, resulting in a lower F1 score. 

It can be observed that many competing methods may exhibit high accuracy but low recall. Recall, or the true positive rate, quantifies the proportion of correctly identified lane pixels relative to all ground truth lane pixels. A low recall indicates that, while the model may accurately classify most background pixels, it struggles to detect lanes, often misclassifying lane pixels as background. The presence of high specificity, which represents the true negative rate, further supports this observation, suggesting that a significant number of pixels are correctly identified as background. In conclusion, a model exhibiting high accuracy but low precision and recall, coupled with high specificity, may not perform effectively in our use case, as it fails to reliably detect lanes.

Overall, the best mIoU, Accuracy and F1 Score indicates that RaceGAN is the most well-rounded algorithm. It also achieves the second best recall value, along with strong precision and specificity. However, there is potential for further improvement, particularly in enhancing both precision and recall.


\subsection{Track Detection Examples under Different Scenarios}


Next, in Figure \ref{fig:superimposed_key}, we present an example of (a) the original input image to our model, (b) the predicted lane mask, and (c) a superimposed version of the predicted mask on the original image, with the red section highlighting the detected lane. In our qualitative assessment of the performance of both our model and the competing models, we use figures similar to Figure \ref{fig:superimposed_key} (c) to visually assess their accuracy and effectiveness.

In Figure \ref{fig:qualitative}, a qualitative analysis of RaceGAN and the competing methods is presented. Due to space constraints, we limit this comparison to the best five of the eight methods in Table \ref{tab:comparison_methods}. The figure includes examples for each of the data categories: normal, dazzle light, color imbalance (green hue, underexposure), curved road, and blurry images. It is evident again that the proposed RaceGAN outperforms the competing methods. For example, Sam2-Unet fails in all samples, incorrectly identifying the race track lane as background. On the other hand, Deeplabv3 suffers from the opposite problem, often identifying background portions as race tracks. WeBACNN and Reseg also suffer from the same problem. This is to be expected, as these lane detection algorithms are not built with optimization for road circuits as the target, so they suffer even when trained with the RoRaTrack dataset.

Visual inspection suggests that among competing methods, Hetnet produces the most accurate predictions, correctly detecting the lane, which is consistent with our findings in Table \ref{tab:comparison_methods}. Still, in the color imbalance (underexposed) and curved road examples, Hetnet appears to mistakenly identify portions of the grass as lane and fails to detect the entire lane. This issue is also observed in the example of color imbalance (green hue), where the lane is not fully detected. In contrast, RaceGAN is able to accurately segment the lane, even reconstructing detailed regions around the vehicle's wheel and chassis. Additionally, it does not incorrectly classify any section of the background as lane, across all the diverse racing scenarios. Our visual observations align with the data presented in Table \ref{tab:comparison_methods} for all methods.

The results presented above demonstrate that although competing methods have shown considerable success in lane detection for traffic datasets, they encounter difficulties when applied to road circuit racing scenarios. The data distribution for racing environments differs significantly, and the unique properties and challenges of these scenarios, as highlighted earlier, contribute to the failure of even the most advanced lane detection models in handling RoRaTrack racing data. We expect that with the availability of RoRaTrack as an open dataset, future algorithms that specifically target road circuits will achieve even better performance. 


\subsection{Computational Cost}

In track detection algorithms, the speed of the algorithm is a critical consideration, as even the best-performing track detection algorithm can be ineffective in a real-time scenario if it cannot process frames quickly enough. Another important point to consider is the computational power needed to run these algorithms. Given that modern racing cars are equipped with highly powerful computing units, there is an increasing need to balance performance with efficiency. Since our track detection algorithm is meant to be run on a resource-constrained processor, our goal is to develop a model that has low floating point operations per second (FLOPs), low memory requirement, and fast inference time. 

To this end, in Table \ref{table:complexity}, we present three values \textemdash FLOPs, the number of parameters in the deep learning model, and inference times. The first two parameters are useful indicators of the processing power required by an algorithm. The inference time indicates how fast one such algorithm can deliver a result in our test environment.

In the table, along with the competing methods, we also present the values for YOLOv8, the model that we used to aid our annotation. YOLOv8 has a large value for FLOPs (12.10G), and a high inference time of 5.15 ms, highlighting one of the primary limitations of using this baseline model for track detection: its substantial computational requirements and relatively slow inference time. Fast inference is crucial for track detection, particularly when the processor is handling data from multiple sensors beyond just cameras. Moreover, we expect the model to perform track detection in real time, which further emphasizes the need for a more efficient solution. Unfortunately, Hetnet, the second best method in Table \ref{tab:comparison_methods}, performs even more poorly \textemdash having an inference time of 15.8 ms \textemdash ruling it out as a viable method for detecting lanes in the RoRaTrack dataset. Ultrafast, Polylanenet, and Samnet also suffer from the same problem. 

Among the remaining methods, Deeplabv3 has the lowest FLOPs with 3.82G FLOPs, WeBACNN exhibits the lowest number of parameters, with 1.15M parameters and an inference time of 0.58 ms, which is the fastest among the competing methods. Our model ranks second in terms of the number of parameters, with 1.59M parameters, and third in terms of inference time, achieving 1.82 ms per frame. This is expected as our model incorporates WeBACNN as a subsystem. Considering the higher mIoU values and the slightly increased inference time, there is a trade-off between a model's accuracy and its computational requirements. Although our model is slower than the fastest method, we believe that its superior performance in terms of mIoU, accuracy, recall, and F1 score justifies the trade-off for more accurate predictions. 

\begin{table}[t]
    \centering
    \caption{FLOPs, Number of Parameters, and Inference Times for different track detection methods.}
    \begin{tabular}{
        >{\columncolor[HTML]{CFE2F3}}c  
        c
        c
        c
    }
        \toprule
        \rowcolor{gray!30}  
        \textbf{Methods} & \textbf{FLOPs (G)} & \textbf{Params (M)} & \textbf{Inf. Time (ms)} \\
        \midrule
        WeBACNN & 7.37 & 1.15 & 0.58 \\
        Ultrafast & 13.26 & 36.51 & 8.4 \\
        Segnet & 64.49 & 29.46 & 1.54 \\
        Reseg & - & - & 2 \\
        Polylanenet & 0.039 & 6.522 & 8.53 \\
        Hetnet & - & - & 15.8 \\
        Deeplabv3 & 3.82 & 12.69 & 1.36 \\
        Samnet & 4.98 & 216.53 & 14.80 \\
        Yolov8 & 12.10 & 3.26 & 5.15 \\
        \textbf{RaceGAN} &\textbf{7.47}& \textbf{1.59}& \textbf{1.82}\\ 
        \rowcolor{gray!10}
        \bottomrule
    \end{tabular}
    \vspace{-15pt}
    \label{table:complexity}
\end{table}

\section{Conclusion}\label{conclusion}

In this work, we address the limited availability of road racing datasets by introducing the RoRaTrack dataset, an open source resource designed for research purposes. The RoRaTrack dataset encompasses a wide variety of common racing scenarios and challenges, providing a comprehensive and valuable tool for researchers to develop models specifically tailored to the complexities of road racing. This is particularly significant, as existing models trained on traffic lane data often struggle when applied to racing environments, underscoring the need for specialized solutions. Building on this foundation, we propose RaceGAN, a novel GAN-based approach for track detection. RaceGAN outperforms current state-of-the-art methods, effectively overcoming the unique challenges of racing data and setting a new benchmark for track detection in autonomous racing.

\section{Acknowledgements}

We acknowledge the valuable contributions of all members of the Black and Gold Racing team at Purdue University for their assistance in data collection. In particular, we express our heartfelt gratitude to Andres Hoyos, Alec Pannunzio, Haoguang Yang, Jiqian Dong, Sashank Modali, Manuel Mar, Zainab Saka and Richard Ajagu for their dedicated efforts. This work was partly supported by Purdue University's Center for Connected and Automated Transportation (CCAT), a part of the larger CCAT consortium, a USDOT Region 5 University Transportation Center funded by the US DOT Award $\#69A3551747105$.

\ifCLASSOPTIONcaptionsoff
  \newpage
\fi
\bibliographystyle{IEEEtran} 

\bibliography{main}

\begin{thebibliography}{10}
\providecommand{\url}[1]{#1}
\csname url@samestyle\endcsname
\providecommand{\newblock}{\relax}
\providecommand{\bibinfo}[2]{#2}
\providecommand{\BIBentrySTDinterwordspacing}{\spaceskip=0pt\relax}
\providecommand{\BIBentryALTinterwordstretchfactor}{4}
\providecommand{\BIBentryALTinterwordspacing}{\spaceskip=\fontdimen2\font plus
\BIBentryALTinterwordstretchfactor\fontdimen3\font minus \fontdimen4\font\relax}
\providecommand{\BIBforeignlanguage}[2]{{%
\expandafter\ifx\csname l@#1\endcsname\relax
\typeout{** WARNING: IEEEtran.bst: No hyphenation pattern has been}%
\typeout{** loaded for the language `#1'. Using the pattern for}%
\typeout{** the default language instead.}%
\else
\language=\csname l@#1\endcsname
\fi
#2}}
\providecommand{\BIBdecl}{\relax}
\BIBdecl

\bibitem{nuscenes}
H.~Caesar, V.~Bankiti, A.~H. Lang, S.~Vora, V.~E. Liong, Q.~Xu, A.~Krishnan, Y.~Pan, G.~Baldan, and O.~Beijbom, ``nuscenes: A multimodal dataset for autonomous driving,'' in \emph{Proceedings of the IEEE/CVF Conference on Computer Vision and Pattern Recognition (CVPR)}, June 2020.

\bibitem{ghosh2024weighted}
S.~Ghosh, Y.-H. Chen, C.-H. Huang, A.~S. M.~M. Jameel, A.~El~Gamal, and S.~Labi, ``Weighted branch aggregation based deep learning model for track detection in autonomous racing,'' in \emph{The Second Tiny Papers Track at ICLR 2024}, 2024.

\bibitem{racecar}
\BIBentryALTinterwordspacing
A.~Kulkarni, J.~Chrosniak, E.~Ducote, F.~Sauerbeck, A.~Saba, U.~Chirimar, J.~Link, M.~Behl, and M.~Cellina, ``Racecar - the dataset for high-speed autonomous racing,'' in \emph{2023 IEEE/RSJ International Conference on Intelligent Robots and Systems (IROS)}.\hskip 1em plus 0.5em minus 0.4em\relax IEEE, Oct. 2023. [Online]. Available: \url{http://dx.doi.org/10.1109/IROS55552.2023.10342053}
\BIBentrySTDinterwordspacing

\bibitem{gta5}
\BIBentryALTinterwordspacing
S.~R. Richter, V.~Vineet, S.~Roth, and V.~Koltun, ``Playing for data: Ground truth from computer games,'' \emph{CoRR}, vol. abs/1608.02192, 2016. [Online]. Available: \url{http://arxiv.org/abs/1608.02192}
\BIBentrySTDinterwordspacing

\bibitem{synthia}
G.~Ros, L.~Sellart, J.~Materzynska, D.~Vazquez, and A.~M. Lopez, ``The synthia dataset: A large collection of synthetic images for semantic segmentation of urban scenes,'' in \emph{Proceedings of the IEEE Conference on Computer Vision and Pattern Recognition (CVPR)}, June 2016.

\bibitem{TuSimple}
\BIBentryALTinterwordspacing
S.~Yoo, H.~Lee, H.~Myeong, S.~Yun, H.~Park, J.~Cho, and D.~H. Kim, ``End-to-end lane marker detection via row-wise classification,'' \emph{CoRR}, vol. abs/2005.08630, 2020. [Online]. Available: \url{https://arxiv.org/abs/2005.08630}
\BIBentrySTDinterwordspacing

\bibitem{CULane}
\BIBentryALTinterwordspacing
X.~Pan, J.~Shi, P.~Luo, X.~Wang, and X.~Tang, ``Spatial as deep: Spatial {CNN} for traffic scene understanding,'' \emph{CoRR}, vol. abs/1712.06080, 2017. [Online]. Available: \url{http://arxiv.org/abs/1712.06080}
\BIBentrySTDinterwordspacing

\bibitem{llamas}
K.~Behrendt and R.~Soussan, ``Unsupervised labeled lane markers using maps,'' in \emph{Proceedings of the IEEE International Conference on Computer Vision}, 2019.

\bibitem{bdd100k}
\BIBentryALTinterwordspacing
F.~Yu, W.~Xian, Y.~Chen, F.~Liu, M.~Liao, V.~Madhavan, and T.~Darrell, ``{BDD100K:} {A} diverse driving video database with scalable annotation tooling,'' \emph{CoRR}, vol. abs/1805.04687, 2018. [Online]. Available: \url{http://arxiv.org/abs/1805.04687}
\BIBentrySTDinterwordspacing

\bibitem{a2d2}
\BIBentryALTinterwordspacing
J.~Geyer, Y.~Kassahun, M.~Mahmudi, X.~Ricou, R.~Durgesh, A.~S. Chung, L.~Hauswald, V.~H. Pham, M.~M{\"{u}}hlegg, S.~Dorn, T.~Fernandez, M.~J{\"{a}}nicke, S.~Mirashi, C.~Savani, M.~Sturm, O.~Vorobiov, M.~Oelker, S.~Garreis, and P.~Schuberth, ``{A2D2:} audi autonomous driving dataset,'' \emph{CoRR}, vol. abs/2004.06320, 2020. [Online]. Available: \url{https://arxiv.org/abs/2004.06320}
\BIBentrySTDinterwordspacing

\bibitem{vil100}
\BIBentryALTinterwordspacing
Y.~Zhang, L.~Zhu, W.~Feng, H.~Fu, M.~Wang, Q.~Li, C.~Li, and S.~Wang, ``{VIL-100:} {A} new dataset and {A} baseline model for video instance lane detection,'' \emph{CoRR}, vol. abs/2108.08482, 2021. [Online]. Available: \url{https://arxiv.org/abs/2108.08482}
\BIBentrySTDinterwordspacing

\bibitem{yolov8}
R.~Varghese and M.~Sambath, ``Yolov8: A novel object detection algorithm with enhanced performance and robustness,'' in \emph{2024 International Conference on Advances in Data Engineering and Intelligent Computing Systems (ADICS)}.\hskip 1em plus 0.5em minus 0.4em\relax IEEE, 2024, pp. 1--6.

\bibitem{hetnet}
\BIBentryALTinterwordspacing
R.~He, J.~Lin, and R.~W.~H. Lau, ``Efficient mirror detection via multi-level heterogeneous learning,'' 2022. [Online]. Available: \url{https://arxiv.org/abs/2211.15644}
\BIBentrySTDinterwordspacing

\bibitem{sam2-unet}
X.~Xiong, Z.~Wu, S.~Tan, W.~Li, F.~Tang, Y.~Chen, S.~Li, J.~Ma, and G.~Li, ``Sam2-unet: Segment anything 2 makes strong encoder for natural and medical image segmentation,'' \emph{arXiv preprint arXiv:2408.08870}, 2024.

\bibitem{sam2}
\BIBentryALTinterwordspacing
N.~Ravi, V.~Gabeur, Y.-T. Hu, R.~Hu, C.~Ryali, T.~Ma, H.~Khedr, R.~R{\"a}dle, C.~Rolland, L.~Gustafson, E.~Mintun, J.~Pan, K.~V. Alwala, N.~Carion, C.-Y. Wu, R.~Girshick, P.~Doll{\'a}r, and C.~Feichtenhofer, ``Sam 2: Segment anything in images and videos,'' \emph{arXiv preprint arXiv:2408.00714}, 2024. [Online]. Available: \url{https://arxiv.org/abs/2408.00714}
\BIBentrySTDinterwordspacing

\bibitem{reseg}
\BIBentryALTinterwordspacing
F.~Visin, M.~Ciccone, A.~Romero, K.~Kastner, K.~Cho, Y.~Bengio, M.~Matteucci, and A.~Courville, ``Reseg: A recurrent neural network-based model for semantic segmentation,'' 2016. [Online]. Available: \url{https://arxiv.org/abs/1511.07053}
\BIBentrySTDinterwordspacing

\bibitem{renet}
\BIBentryALTinterwordspacing
F.~Visin, K.~Kastner, K.~Cho, M.~Matteucci, A.~Courville, and Y.~Bengio, ``Renet: A recurrent neural network based alternative to convolutional networks,'' 2015. [Online]. Available: \url{https://arxiv.org/abs/1505.00393}
\BIBentrySTDinterwordspacing

\bibitem{deeplabv3}
\BIBentryALTinterwordspacing
L.-C. Chen, G.~Papandreou, F.~Schroff, and H.~Adam, ``Rethinking atrous convolution for semantic image segmentation,'' 2017. [Online]. Available: \url{https://arxiv.org/abs/1706.05587}
\BIBentrySTDinterwordspacing

\bibitem{segnet}
V.~Badrinarayanan, A.~Kendall, and R.~Cipolla, ``Segnet: A deep convolutional encoder-decoder architecture for image segmentation,'' \emph{IEEE Transactions on Pattern Analysis and Machine Intelligence}, vol.~39, no.~12, pp. 2481--2495, 2017.

\bibitem{vgg16}
\BIBentryALTinterwordspacing
K.~Simonyan and A.~Zisserman, ``Very deep convolutional networks for large-scale image recognition,'' 2015. [Online]. Available: \url{https://arxiv.org/abs/1409.1556}
\BIBentrySTDinterwordspacing

\bibitem{ultrafast}
Z.~Qin, H.~Wang, and X.~Li, ``Ultra fast structure-aware deep lane detection,'' in \emph{Computer Vision--ECCV 2020: 16th European Conference, Glasgow, UK, August 23--28, 2020, Proceedings, Part XXIV 16}.\hskip 1em plus 0.5em minus 0.4em\relax Springer, 2020, pp. 276--291.

\bibitem{Feng_2022_CVPR}
Z.~Feng, S.~Guo, X.~Tan, K.~Xu, M.~Wang, and L.~Ma, ``Rethinking efficient lane detection via curve modeling,'' in \emph{Proceedings of the IEEE/CVF Conference on Computer Vision and Pattern Recognition (CVPR)}, June 2022, pp. 17\,062--17\,070.

\bibitem{poly}
\BIBentryALTinterwordspacing
L.~T. Torres, R.~F. Berriel, T.~M. Paix{\~{a}}o, C.~Badue, A.~F.~D. Souza, and T.~Oliveira{-}Santos, ``Polylanenet: Lane estimation via deep polynomial regression,'' \emph{CoRR}, vol. abs/2004.10924, 2020. [Online]. Available: \url{https://arxiv.org/abs/2004.10924}
\BIBentrySTDinterwordspacing

\bibitem{linecnn}
X.~Li, J.~Li, X.~Hu, and J.~Yang, ``Line-cnn: End-to-end traffic line detection with line proposal unit,'' \emph{IEEE Transactions on Intelligent Transportation Systems}, vol.~21, no.~1, pp. 248--258, 2020.

\bibitem{laneatt}
L.~Tabelini, R.~Berriel, T.~M. Paixao, C.~Badue, A.~F. De~Souza, and T.~Oliveira-Santos, ``Keep your eyes on the lane: Real-time attention-guided lane detection,'' in \emph{Proceedings of the IEEE/CVF Conference on Computer Vision and Pattern Recognition (CVPR)}, June 2021, pp. 294--302.

\bibitem{Zheng_2022_CVPR}
T.~Zheng, Y.~Huang, Y.~Liu, W.~Tang, Z.~Yang, D.~Cai, and X.~He, ``Clrnet: Cross layer refinement network for lane detection,'' in \emph{Proceedings of the IEEE/CVF Conference on Computer Vision and Pattern Recognition (CVPR)}, June 2022, pp. 898--907.

\bibitem{Wang_2022_CVPR}
J.~Wang, Y.~Ma, S.~Huang, T.~Hui, F.~Wang, C.~Qian, and T.~Zhang, ``A keypoint-based global association network for lane detection,'' in \emph{Proceedings of the IEEE/CVF Conference on Computer Vision and Pattern Recognition (CVPR)}, June 2022, pp. 1392--1401.

\bibitem{cyclegan}
J.-Y. Zhu, T.~Park, P.~Isola, and A.~A. Efros, ``Unpaired image-to-image translation using cycle-consistent adversarial networks,'' in \emph{2017 IEEE International Conference on Computer Vision (ICCV)}, 2017, pp. 2242--2251.

\bibitem{wgan}
\BIBentryALTinterwordspacing
M.~Arjovsky, S.~Chintala, and L.~Bottou, ``Wasserstein gan,'' 2017. [Online]. Available: \url{https://arxiv.org/abs/1701.07875}
\BIBentrySTDinterwordspacing

\end{thebibliography}

\end{document}